\documentclass[letterpaper]{article}
\usepackage[preprint]{lakemlb}

\usepackage[hyphens]{url} 
\usepackage{graphicx} 
\usepackage{natbib} 
\usepackage{caption} 
\usepackage{booktabs}
\usepackage{multirow}
\usepackage{amsmath}
\usepackage{amsfonts}
\usepackage{amssymb}
\usepackage{xspace}
\usepackage{amsthm}
\usepackage{makecell}
\usepackage{comment}
\usepackage{subcaption}

\theoremstyle{definition}
\newtheorem{example}{Example}

\title{LakeMLB: Data Lake Machine Learning Benchmark}

\author{
Feiyu Pan,\textsuperscript{\rm 1}
Tianbin Zhang,\textsuperscript{\rm 1}
Aoqian Zhang,\textsuperscript{\rm 2}
Yu Sun,\textsuperscript{\rm 3}\\
Zheng Wang,\textsuperscript{\rm 1}
Lixing Chen,\textsuperscript{\rm 1}
Li Pan,\textsuperscript{\rm 1}
Jianhua Li\textsuperscript{\rm 1}
}
\affiliations{
\textsuperscript{\rm 1}Shanghai Jiao Tong University, Shanghai, China\\
\textsuperscript{\rm 2}Beijing Institute of Technology, Beijing, China\\
\textsuperscript{\rm 3}Nankai University, Tianjin, China\\
\textsuperscript{\rm 1}\texttt{\{feiyupan,tianbinzhang,wzheng,lxchen,panli,lijh888\}@sjtu.edu.cn}\\
\textsuperscript{\rm 2}\texttt{aoqian.zhang@bit.edu.cn}
\quad
\textsuperscript{\rm 3}\texttt{sunyu@nankai.edu.cn}
}

\begin{document}

\maketitle

% ...
\begin{abstract}
Data lakes have become a fundamental platform for large-scale machine learning by enabling flexible management of heterogeneous data.
Despite their growing importance, standardized benchmarks for evaluating machine learning performance in data lake environments remain scarce.
To address this gap, we present \textbf{LakeMLB} (Data \textbf{Lake} \textbf{M}achine \textbf{L}earning \textbf{B}enchmark), the first benchmark designed for multi-table machine learning in data lakes. LakeMLB focuses on two representative scenarios, \textit{Union} and \textit{Join}, and provides six real-world datasets spanning diverse domains. 
It supports three representative multi-table learning paradigms: pre-training, data augmentation, and feature augmentation, together with standardized data splits and evaluation protocols.
We conduct extensive experiments with state-of-the-art tabular learning methods and provide insights into their performance across different data lake scenarios.
We release both datasets and code to facilitate rigorous research on machine learning in data lake ecosystems; the benchmark is available at \url{https://github.com/zhengwang100/LakeMLB}.
\end{abstract}


\section{Introduction}\label{sec:introduction}

The exponential growth of data in both volume and variety has made data lakes~\cite{sawadogo2021data} a fundamental infrastructure for modern data management. Unlike traditional data warehouses, data lakes store structured, semi-structured, and unstructured data in raw or near-raw formats on scalable object storage, providing high flexibility for large-scale data analysis. More recently, the Lakehouse architecture~\cite{armbrust2021lakehouse} has combined the openness of data lakes with the reliability of data warehouses by introducing ACID transactions, schema enforcement, and metadata management. This evolution transforms heterogeneous data collections into analyzable table-centric repositories, making data lakes an increasingly attractive platform for enterprise AI applications, including large-scale model training, real-time analytics, and reproducible machine learning.

\begin{figure}[t]
\centering
\includegraphics[width=\linewidth]{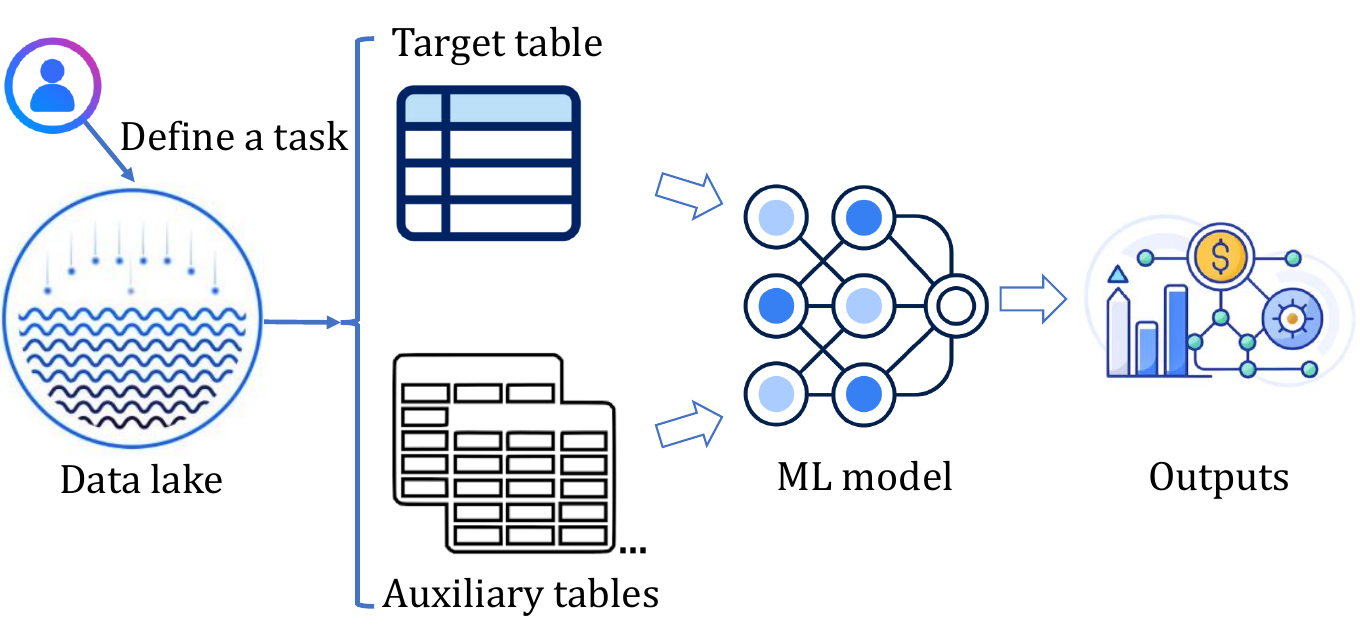}
\caption{Machine learning pipeline in data lakes.}
\label{fig:overview}
\end{figure}

As illustrated in Figure~\ref{fig:overview}, machine learning workflows in data lake environments typically comprise three stages.
The first is \textit{task definition}, which specifies the prediction target and identifies the target table, guiding subsequent data preparation strategies and evaluation metrics. 
The second is \textit{related table discovery}, which identifies auxiliary tables relevant to the target to enrich contextual information. 
The third is \textit{model learning}, which leverages all identified tables to train and fine-tune machine learning models, ultimately producing predictions for the defined tasks. 
Overall, data lakes facilitate machine learning by discovering, integrating, and jointly modeling heterogeneous tables from multiple sources, making multi-table learning a fundamental capability of modern data-centric AI systems.

Despite the growing adoption of data lakes for AI, standardized benchmarks for evaluating the \emph{model learning} stage remain largely absent. Existing data lake benchmarks, such as LakeBench~\cite{srinivas2023lakebench,deng2024lakebench}, primarily focus on table discovery rather than downstream learning. Meanwhile, traditional tabular learning benchmarks, such as the UCI Machine Learning Repository~\cite{asuncion2007uci} and OpenML Benchmarking Suites~\cite{bischl2openml}, are limited to single-table tasks. Although recent multi-table benchmarks, including RelBench~\cite{robinson2024relbench} and SJTUTables~\cite{li2024rllm}, consider relational data, they are designed for conventional database settings and do not capture the heterogeneous data organization and flexible integration patterns characteristic of modern data lakes.

\begin{figure}[t]
\centering
\includegraphics[width=\linewidth]{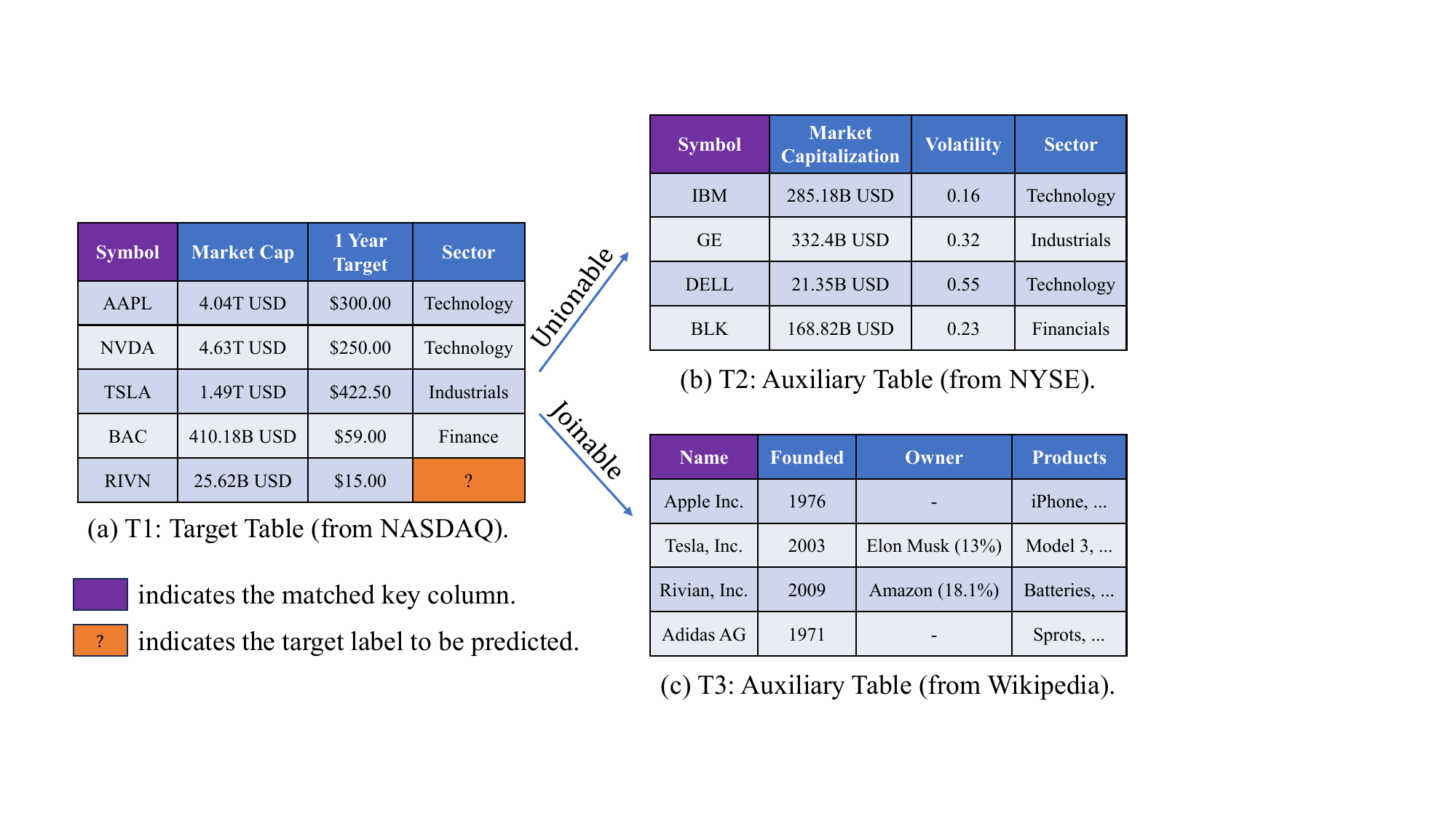} 
\caption{Illustration of unionable and joinable tables.}
\label{fig_data_example}
\end{figure}


To fill this gap, we introduce \textbf{LakeMLB} (Data \textbf{Lake} \textbf{M}achine \textbf{L}earning \textbf{B}enchmark), designed to evaluate machine learning performance in data lake environments. As shown in Figure~\ref{fig_data_example}, LakeMLB focuses on two representative multi-table scenarios, \textit{Union} and \textit{Join}. For each scenario, we construct three real-world datasets spanning diverse domains, including government open data, financial markets, Wikipedia, music platforms, and online marketplaces. 
The \textit{Union} datasets capture semantic inconsistencies and distribution shifts across heterogeneous sources, whereas the \textit{Join} datasets enrich target tables with complementary information through fuzzy entity matching.
All datasets adopt standardized train/validation/test splits and balanced row-level prediction tasks to ensure fair and reproducible evaluation.


Furthermore, LakeMLB evaluates three representative multi-table integration paradigms: \textit{pre-training}, \textit{data augmentation}, and \textit{feature augmentation}. Based on extensive experiments with both classical machine learning and modern deep tabular models, we obtain three key findings: (1) Auxiliary table utilization is highly scenario-dependent: pre-training performs best for \textit{Union} tasks, whereas feature augmentation is most effective for \textit{Join} tasks. (2) Transfer learning provides a robust paradigm for multi-table learning, consistently achieving strong performance across both scenarios. (3) Existing tabular foundation models do not fully exploit auxiliary tables, despite their strong single-table performance, highlighting a promising direction for future research, particularly by leveraging their strong textual understanding capabilities.

The main contributions of this paper are summarized as follows:
\begin{enumerate}
    \item We introduce LakeMLB, the first comprehensive benchmark for evaluating multi-table machine learning tasks in data lake environments.
    \item We conduct extensive evaluation and analysis of state-of-the-art tabular learning methods on this benchmark, providing insights into their performance across complex multi-table scenarios.
    \item We release the benchmark datasets and code as a community resource to advance research on machine learning within data lake ecosystems.
\end{enumerate}


\section{Preliminaries and Task Setting}\label{sec:preliminaries}
In this section, we formally introduce the notions of \emph{table unionability} and \emph{table joinability}, which frequently arise in data lake environments, and define the corresponding machine learning task that leverages these two types of auxiliary tables.

\subsection{Table Unionability}
Following standard definitions in the literature~\cite{nargesian2018table,khatiwada2023santos}, two tables $A$ and $B$ are said to be \emph{unionable} if a subset of their columns can be aligned and subsequently merged.
More specifically, a column $c_1$ from table $A$ is unionable with a column $c_2$ from table $B$ if their values are drawn from the same underlying domain.
Unionable tables are commonly used to vertically extend a target table by appending additional rows, thereby increasing the number of training samples and improving the performance of downstream machine learning models.

\begin{example}[Illustration of Table Unionability]
As shown in Figure~\ref{fig_data_example}, tables $T_1$ and $T_2$ are unionable because many of their columns either exactly match or convey equivalent semantics.
For instance, the column \emph{Symbol} in $T_1$ and the column \emph{Symbol} in $T_2$ are identical, while the column \emph{Market Cap} in $T_1$ and the column \emph{Market Capitalization} in $T_2$ are syntactically different but semantically equivalent.
Intuitively, by aligning these unionable columns, the two tables can be vertically concatenated to form a larger table with additional rows, which provides more training samples for downstream machine learning tasks.
\end{example}

\begin{table*}[t]
\centering
\renewcommand{\arraystretch}{1.2}
{\small
\begin{tabular}{l l c c c c c c c c c}
\toprule
\multirow{2}{*}{\textbf{Relation}}
& \multirow{2}{*}{\textbf{Dataset}}
& \multirow{2}{*}{\textbf{Task}}
& \multicolumn{4}{c}{\textbf{Target Table}}
& \multicolumn{4}{c}{\textbf{Auxiliary Table}} \\
\cmidrule(lr){4-7} \cmidrule(lr){8-11}
&
&
& \textbf{Name} & \textbf{\#row} & \textbf{\#col} & \textbf{\#class}
& \textbf{Name} & \textbf{\#row} & \textbf{\#col} & \textbf{\#class} \\
\midrule
\multirow{3}{*}{\textbf{Union}}
& MSTraffic & collision-type
& Maryland & 10,800 & 37 & 9
& Seattle & 10,800 & 50 & 9 \\
& NCBuilding & violation-category
& NewYork & 30,000 & 40 & 30
& Chicago & 37,000 & 23 & 37 \\
& NCTaxi & dropoff-location
& NewYork & 100,000 & 19 & 50
& Chicago & 100,000 & 19 & 50 \\
\midrule
\multirow{3}{*}{\textbf{Join}}
& NNStocks & sector-cls
& NNList & 1,078 & 11 & 11
& NNWiki & 937 & 22 & -- \\
& DSMusic & music-genre
& Discogs & 11,000 & 5 & 11
& Spotify & 11,000 & 21 & -- \\
& AGBooks & book-category
& Amazon & 100,000 & 11 & 40
& Goodreads & 100,000 & 20 & -- \\
\bottomrule
\end{tabular}
}
\caption{Benchmark datasets statistics.}
\label{tab:datasets}
\end{table*}

\subsection{Table Joinability}
Following standard definitions in the literature~\cite{zhu2019josie,srinivas2023lakebench}, two columns $c_1$ from table $A$ and $c_2$ from table $B$ are considered \emph{joinable} if they correspond to the same semantic type and exhibit overlapping values.
In practice, joinable tables are primarily used to horizontally enrich a target table by appending additional attribute columns, which often improves downstream machine learning performance.

\begin{example}[Illustration of Table Joinability]
As illustrated in Figure~\ref{fig_data_example}, tables $T_1$ and $T_3$ are joinable because their first columns convey highly similar semantics and share overlapping values.
Specifically, (1) the column names are semantically related (e.g., the column \emph{Symbol} in $T_1$ and the column \emph{Name} in $T_3$ both denote company identifiers), and
(2) the corresponding cell values exhibit fuzzy overlaps (e.g., \texttt{AAPL} in $T_1$ and \texttt{Apple Inc.} in $T_3$).
Intuitively, by joining the two tables on these columns, additional attributes from $T_3$ can be appended to $T_1$, enabling feature enrichment for downstream machine learning tasks.
\end{example}

\subsection{Machine Learning in Data Lakes}\label{preliminaries:problem_definition}
In general, a machine learning task aims to learn a model that predicts the label of a given instance.
In the tabular setting, we broadly define such a task as predicting the value of a specific cell (i.e., the label) for a given row (i.e., an instance) in a table\footnote{For simplicity, we focus on predicting values for existing rows. Extending this formulation to temporal forecasting settings is left for future work.}.

In a data lake environment, a user first specifies a prediction task by selecting a target table and a particular column whose values are to be predicted.
Given the target table, the data lake system retrieves a set of auxiliary tables through table discovery, which may include joinable tables, unionable tables, or both.
For simplicity, we assume the presence of a single auxiliary table.
The objective is to jointly leverage the target table and the auxiliary table to train a machine learning model that achieves improved predictive performance on the target table.

\begin{example}[Illustrative Machine Learning Task in a Data Lake]
As shown in Figure~\ref{fig_data_example}, a user aims to predict the value of the column \emph{Sector} for the fifth row of table $T_1$ (sourced from the NASDAQ website), corresponding to the stock symbol \texttt{RIVN}.
Through table discovery in the data lake, two related tables are identified: a unionable table $T_2$ and a joinable table $T_3$.
The goal is to exploit these auxiliary tables (i.e., $T_2$ and $T_3$) to enhance the performance of the machine learning model on the prediction task over $T_1$.
Intuitively, this can be achieved by either increasing the number of training samples via table union or enriching feature representations via table join.
\end{example}

\section{Benchmark Design}
\label{sec:benchmark}
\textbf{LakeMLB} (Data \textbf{Lake} \textbf{M}achine \textbf{L}earning \textbf{B}enchmark) is a benchmark suite for evaluating machine learning models in data lake environments, with an emphasis on two fundamental paradigms of multi-source tabular integration: \emph{Union} and \emph{Join}. 
For simplicity, each benchmark instance contains a target table paired with only one auxiliary table. 
To ensure standardized and fair evaluation, all datasets are constructed with balanced class distributions. Target tables are split into training ($70\%$), validation ($10\%$), and test ($20\%$) sets. 
Table~\ref{tab:datasets} summarizes the detailed statistics of all datasets.
Additional details are provided in Appendix~\ref{app:benchmark}.

\subsection{Union Case}
The union case comprises three datasets spanning diverse application domains.
In each dataset, the target and auxiliary tables originate from the same domain but are collected from different data sources.
Although they share similar label spaces, the tables exhibit semantic inconsistencies and distributional shifts, making effective knowledge transfer non-trivial.

\noindent \subsubsection*{\textbf{MSTraffic} (\textbf{M}aryland and \textbf{S}eattle \textbf{Traffic} Collisions)}
This dataset is constructed from publicly available U.S. government data and focuses on urban traffic accident analysis.
The target table (\textit{Maryland}) is derived from traffic safety records released by the State of Maryland and contains detailed contextual information at the time of each accident, such as weather conditions and road environments.
The prediction task is to identify the \textit{collision type}.
The auxiliary table (\textit{Seattle}) is collected from traffic collision data published by the Seattle Department of Transportation.
While it also includes accident-type labels, its categorization scheme only partially overlaps with that of the target table, resulting in semantic mismatches.

\noindent \subsubsection*{\textbf{NCBuilding} (\textbf{N}ew York and \textbf{C}hicago \textbf{Building} Violations)}
This dataset is sourced from open U.S. government platforms and focuses on urban building violation complaints.
The target table (\textit{New\allowbreak York}) is obtained from New York City housing maintenance violation records, which document inspection outcomes and detailed violation descriptions.
The task is to predict the \textit{violation category}.
The auxiliary table (\textit{Chicago}) is drawn from building violation records released by the City of Chicago.
Despite belonging to the same application domain, the violation definitions and labeling standards differ from those of the target table.

\noindent \subsubsection*{\textbf{NCTaxi} (\textbf{N}ew York and \textbf{C}hicago \textbf{Taxi} Trips)}
This dataset is constructed from taxi trip records sourced from NYC Open Data and the Chicago Data Portal and is designed for destination-zone classification of taxi trips.
The target table (\textit{New\allowbreak York}) is derived from the 2023 New York City Yellow Taxi Trip Records and contains temporal, spatial, and fare-related attributes for each trip.
The task is to predict the trip's drop-off \textit{Taxi Zone}.
The auxiliary table (\textit{Chicago}) is derived from the 2023 Chicago Taxi Trips dataset and contains comparable trip attributes together with drop-off community area labels.
Although both tables describe urban taxi trips, their destination labels are defined according to distinct city-specific geographic zoning systems, resulting in semantic misalignment and distribution shift between the two tables.

\subsection{Join Case}
The join case consists of three datasets in which the target and auxiliary tables are related but lack explicit foreign key constraints.
To establish table connections in a unified manner, we employ a fuzzy entity matching strategy.
Specifically, candidate linking columns are encoded into textual embeddings, and cosine similarity is used to perform approximate 1-nearest-neighbor (1-NN) retrieval, yielding weak yet practical join relationships.

\noindent \subsubsection*{\textbf{NNStocks} (\textbf{N}ASDAQ and \textbf{N}YSE \textbf{Stocks})}
This dataset focuses on industry classification for publicly listed U.S. companies.
The target table (\textit{NNList}) is constructed from official company listings on the NASDAQ and the New York Stock Exchange.
The task is to predict the company \textit{sector classification}.
The auxiliary table (\textit{NNWiki}) is crawled from corresponding company infoboxes on Wikipedia.
It is linked to the target table via fuzzy matching between company names and Wikipedia page titles, providing enriched semantic attributes.

\noindent \subsubsection*{\textbf{DSMusic} (\textbf{D}iscogs and \textbf{S}potify \textbf{Music})}
This dataset addresses the music genre classification task.
The target table (\textit{Discogs}) is derived from the public Discogs database and contains metadata for individual music tracks.
The task is to predict the \textit{genre category}.
The auxiliary table (\textit{Spotify}) is collected from the music website Spotify, which augments the target data with complementary attributes such as popularity indicators and rhythmic features.
It is linked to the target table via fuzzy matching between track titles and Spotify track names, providing complementary audio and popularity attributes.

\noindent \subsubsection*{\textbf{AGBooks} (\textbf{A}mazon and \textbf{G}oodreads \textbf{Books})}
This dataset is designed for a book category classification task.
The target table (\textit{Amazon}) is derived from the Amazon Reviews 2023 dataset and contains product metadata for books listed on Amazon.
The task is to predict the \textit{category label} of each book.
The auxiliary table (\textit{Goodreads}) is derived from the Goodreads Books dataset and complements the target data with bibliographic and community-derived attributes, including publication information, rating and review statistics, and book descriptions.
The two tables are linked through fuzzy entity matching between Amazon and Goodreads book titles, thereby enriching the target table with complementary book metadata.



\section{Benchmark Discussion}

\subsection{Benchmark Strengths and Limitations}
\label{sec:str_lim}

\subsubsection{\textbf{Strengths.}}
LakeMLB offers three main strengths. \textbf{(1) Novelty.} To the best of our knowledge, LakeMLB is the first benchmark designed specifically for machine learning in data lake environments. Unlike existing benchmarks built on relational databases, it targets two fundamental data lake integration operations, \textit{Union} and \textit{Join}, enabling systematic evaluation of tabular learning models under realistic multi-table settings. \textbf{(2) Simplicity.} To facilitate controlled analysis, LakeMLB restricts each task to integrating two tables and focuses on balanced classification tasks. These design choices reduce confounding factors and allow researchers to isolate the impact of data integration strategies. \textbf{(3) Ease of use.} LakeMLB is lightweight and easy to deploy. Its moderate scale enables experiments on commodity hardware, while a unified training and evaluation pipeline ensures reproducibility and lowers the barrier to adoption.

\subsubsection{\textbf{Limitations.}}
LakeMLB also has several limitations. \textbf{(1) Limited task diversity.} The current benchmark considers only classification and does not cover other important tasks, such as regression and ranking, nor practical challenges including label imbalance and noisy labels. \textbf{(2) Limited scale.} Each sub-dataset contains only two tables, with dataset sizes ranging from approximately 0.8K to 100K rows, making the benchmark smaller than real-world data lakes. \textbf{(3) Limited scenario complexity.} LakeMLB currently focuses on pairwise table integration and does not include more challenging settings involving multiple auxiliary tables, multilingual data, or multimodal tabular data.

\subsection{Task Potential Solutions}
\label{sec:chall_sol}

Although existing tabular learning methods are not explicitly designed for \emph{unionable} or \emph{joinable} tables, they can be naturally adapted as competitive baselines under simplified settings.

\paragraph{Learning with Unionable Tables.} As illustrated in Figure~\ref{fig_data_example}, unionable tables can be regarded as additional training data, despite differences in their feature and label spaces. A natural solution is the \textbf{pre-training} paradigm, where a model is first pre-trained on auxiliary tables and then fine-tuned on the target task. However, inconsistent feature spaces, heterogeneous table semantics, and the difficulty of designing effective pre-training objectives limit the applicability of this approach. Another direction is \textbf{data augmentation}, which retrieves semantically similar samples from auxiliary tables and incorporates them as pseudo-labeled training instances. The effectiveness of this strategy largely depends on pseudo-label quality, as semantic inconsistencies across tables can introduce substantial noise.

\paragraph{Learning with Joinable Tables.} As shown in Figure~\ref{fig_data_example}, joinable tables naturally support a \textbf{feature augmentation} paradigm by enriching each target instance with additional features. A straightforward implementation identifies matching rows in auxiliary tables using semantic keys (e.g., via 1-NN matching) and concatenates the retrieved features with those of the target table before applying standard tabular learning models. The primary challenge lies in accurate row matching, since incorrect joins introduce noisy features that can significantly degrade downstream performance.
\section{Experiments}
\label{sec:experiments}


\begin{table*}[!t]
\centering
\setlength{\tabcolsep}{2.5pt}

{\small
\begin{tabular*}{\textwidth}{
@{\extracolsep{\fill}}
lcccc cccc cccc
@{}
}
\toprule
\multirow{2}{*}{\textbf{Methods}}
& \multicolumn{4}{c}{\textbf{MSTraffic}}
& \multicolumn{4}{c}{\textbf{NCBuilding}}
& \multicolumn{4}{c}{\textbf{NCTaxi}} \\
\cmidrule(lr){2-5}
\cmidrule(lr){6-9}
\cmidrule(lr){10-13}

& $T_{\mathrm{tgt}}$
& $+\mathrm{PT}_{T_{\mathrm{aux}}}$
& $+\mathrm{DA}_{T_{\mathrm{aux}}}$
& $+\mathrm{FA}_{T_{\mathrm{aux}}}$

& $T_{\mathrm{tgt}}$
& $+\mathrm{PT}_{T_{\mathrm{aux}}}$
& $+\mathrm{DA}_{T_{\mathrm{aux}}}$
& $+\mathrm{FA}_{T_{\mathrm{aux}}}$

& $T_{\mathrm{tgt}}$
& $+\mathrm{PT}_{T_{\mathrm{aux}}}$
& $+\mathrm{DA}_{T_{\mathrm{aux}}}$
& $+\mathrm{FA}_{T_{\mathrm{aux}}}$ \\
\midrule
XGBoost
& 41.76 & - & 42.81 & 43.96
& 37.15 & - & 38.26 & 38.45
& 13.07 & - & 11.98 & 12.94 \\
LightGBM
& 41.53 & - & 41.92 & 43.28
& 37.93 & - & 39.24 & 38.89
& 14.57 & - & 12.72 & 14.24 \\
CatBoost
& 41.62 & - & 42.25 & 43.36
& 43.37 & - & 42.56 & 44.22
& 11.14 & - & 13.81 & 10.64 \\
\midrule
TabTransformer
& 41.00 & - & 41.37 & 42.00
& 39.40 & - & 39.42 & 39.63
& 10.25 & - & 8.90 & 8.79 \\
FT-Transformer
& 38.38 & - & 37.63 & 39.75
& 41.27 & - & 40.50 & 41.85
& 18.24 & - & 18.04 & 19.00 \\
SAINT
& 38.40 & - & 37.93 & 39.78
& 41.04 & - & 32.63 & 38.67
& 19.19 & - & 17.46 & 20.01 \\
Trompt
& 36.35 & - & 38.03 & 38.02
& 39.84 & - & 40.30 & 38.17
& 16.99 & - & 15.43 & 16.78 \\
ExcelFormer
& 38.55 & - & 40.05 & 41.31
& 34.71 & - & 37.16 & 37.00
& 20.85 & - & 20.44 & 20.42 \\
\midrule
TransTab
& 35.59 & 35.43 & 34.93 & 34.76
& 39.87 & 40.58 & 38.92 & 38.99
& 16.12 & 16.03 & 16.12 & 15.92 \\
CARTE
& 40.41 & 40.98 & 39.52 & 40.92
& 42.46 & 43.02 & 39.01 & 42.69
& 21.72 & 22.16 & 21.54 & 21.19 \\
\midrule
TabICLv2
& 43.17 & - & 41.46 & 44.77
& 36.64 & - & 34.81 & 36.93
& 21.36 & - & 18.55 & 20.46 \\
TabPFNv3
& 41.51 & - & 41.17 & 42.94
& 37.70 & - & 32.17 & 33.09
& 23.12 & - & 22.47 & 23.15 \\
TabPFNv3-Plus
& 42.50 & - & 54.15 & 42.80
& 51.00 & - & 48.80 & 51.50
& 19.40 & - & 13.40 & 18.50 \\
\bottomrule
\end{tabular*}
}

\caption{Performance comparison on Union-based datasets
(Accuracy, in \%).
``$T_{\mathrm{tgt}}$'' denotes training solely on the target table,
``$+\mathrm{PT}_{T_{\mathrm{aux}}}$'' denotes pre-training using
the auxiliary table,
``$+\mathrm{DA}_{T_{\mathrm{aux}}}$'' represents
pseudo-label-based data augmentation by adding auxiliary samples
to the target table,
and ``$+\mathrm{FA}_{T_{\mathrm{aux}}}$'' corresponds to feature
augmentation via horizontal concatenation of the 1NN-neighbor
rows from the auxiliary table into the target table.}
\label{tab:union_results}
\end{table*}

\subsection{Experimental Setup}
\label{subsec:setup}

\subsubsection*{\textbf{Multi-table Learning Strategies.}}
As outlined in Section~\ref{sec:chall_sol}, in addition to the single-table setting (which trains and evaluates solely on the target table), we consider three representative multi-table strategies that leverage auxiliary tables. These are summarized below; further details are provided in Appendix~\ref{app:aux_strategies}.
\begin{itemize}
    \item \textbf{Pre-training-based Strategy (PT).} Models are pre-trained on auxiliary tables to learn general-purpose representations or parameter initializations, followed by fine-tuning on the target table. This strategy is applicable primarily to models supporting transfer learning.

    \item \textbf{Data Augmentation-based Strategy (DA).} Auxiliary samples are incorporated either via identical or semantically similar labels, or through pseudo-label generation. This approach enriches the training set at the data level.

    \item \textbf{Feature Augmentation-based Strategy (FA).} For each target sample, the most similar row from auxiliary tables is retrieved using $1$-nearest-neighbor (1-NN) search (approximate search for scalability). The retrieved features are concatenated with the target features, effectively expanding the feature space horizontally. 
\end{itemize}

\subsubsection*{\textbf{Evaluation Metrics.}}
Since all LakeMLB tasks are classification problems, we adopt \emph{Accuracy} as the primary metric, following the benchmark's default training, validation, and test splits. For deep models requiring hyperparameter tuning, we conduct grid search based on validation performance. Additional details and hyperparameter settings are provided in Appendix~\ref{app:exp_details}.

\subsection{Compared Methods}
\label{subsec:baselines}

We evaluate four categories of representative and competitive baseline methods, covering both classical and modern tabular learning paradigms:
\begin{itemize}
    \item \textbf{Classical Tree-based Methods.}
    We consider three widely adopted tree-based ensemble models: XGBoost~\cite{Chen2016XGBoost}, CatBoost~\cite{Prokhorenkova2018CatBoost}, and LightGBM~\cite{Ke2017LightGBM}.
    These methods have long been regarded as strong baselines for tabular learning due to their stable training behavior, relatively small model sizes, and robustness to heterogeneous feature types.

    \item \textbf{Single-table Tabular Neural Networks.}
    We evaluate a set of representative deep tabular learning models, including TabTransformer~\cite{huang2020tabtransformer}, FT-Transformer~\cite{gorishniy2021revisiting}, SAINT~\cite{somepalli2021saint}, ExcelFormer~\cite{chen2023excelformer}, and Trompt~\cite{chen2023trompt}.
    These methods are typically built upon Transformer architectures, where categorical features are embedded and high-order feature interactions are modeled via self-attention mechanisms.
    However, they are inherently limited to single-table scenarios.

    \item \textbf{Transfer Learning-based Tabular Models.}
    We include two multi-table deep learning approaches based on transfer learning: TransTab~\cite{wang2022transtab} and CARTE~\cite{kim2024carte}.
    These methods pre-train models on auxiliary tables and subsequently fine-tune them on the target table, enabling cross-table knowledge transfer.

    \item \textbf{Foundation Models for Tabular Data.}
    We also evaluate recently proposed foundation models for tabular data, including TabPFN~\cite{Hollmann2023TabPFN} and TabICL~\cite{qu2025tabicl}.
    These models are typically pre-trained on large-scale synthetic tabular datasets to acquire strong inductive priors and generalization capabilities across diverse downstream tasks. We adopt their in-context learning inference paradigm and use TabPFNv3 and TabICLv2 with default configurations. We additionally evaluate TabPFNv3-Plus, a commercial API-based variant that further supports tabular datasets containing textual columns.
\end{itemize}

\begin{table*}[!t]
\centering
\setlength{\tabcolsep}{2.5pt}

{\small
\begin{tabular*}{\textwidth}{
@{\extracolsep{\fill}}
lcccc cccc cccc
@{}
}
\toprule
\multirow{2}{*}{\textbf{Methods}}
& \multicolumn{4}{c}{\textbf{NNStocks}}
& \multicolumn{4}{c}{\textbf{DSMusic}}
& \multicolumn{4}{c}{\textbf{AGBooks}} \\
\cmidrule(lr){2-5}
\cmidrule(lr){6-9}
\cmidrule(lr){10-13}

& $T_{\mathrm{tgt}}$
& $+\mathrm{PT}_{T_{\mathrm{aux}}}$
& $+\mathrm{DA}_{T_{\mathrm{aux}}}$
& $+\mathrm{FA}_{T_{\mathrm{aux}}}$

& $T_{\mathrm{tgt}}$
& $+\mathrm{PT}_{T_{\mathrm{aux}}}$
& $+\mathrm{DA}_{T_{\mathrm{aux}}}$
& $+\mathrm{FA}_{T_{\mathrm{aux}}}$

& $T_{\mathrm{tgt}}$
& $+\mathrm{PT}_{T_{\mathrm{aux}}}$
& $+\mathrm{DA}_{T_{\mathrm{aux}}}$
& $+\mathrm{FA}_{T_{\mathrm{aux}}}$ \\
\midrule
XGBoost
& 21.21 & - & 21.04 & 24.24
& 36.09 & - & 35.58 & 36.73
& 31.78 & - & 31.20 & 33.06 \\
LightGBM
& 20.35 & - & 19.18 & 26.41
& 34.73 & - & 36.20 & 35.91
& 31.72 & - & 31.36 & 32.94 \\
CatBoost
& 21.65 & - & 21.95 & 23.81
& 32.50 & - & 32.79 & 33.82
& 25.95 & - & 25.25 & 28.56 \\
\midrule
TabTransformer
& 11.69 & - & 11.95 & 20.00
& 25.03 & - & 25.55 & 26.03
& 13.42 & - & 13.36 & 20.65 \\
FT-Transformer
& 10.30 & - & 9.74 & 15.54
& 28.12 & - & 27.84 & 31.51
& 12.48 & - & 12.02 & 21.46 \\
SAINT
& 12.16 & - & 9.05 & 17.27
& 28.97 & - & 26.37 & 30.66
& 11.75 & - & 6.09 & 16.54 \\
Trompt
& 10.52 & - & 11.95 & 15.19
& 28.89 & - & 27.33 & 29.93
& 15.56 & - & 15.68 & 19.76 \\
ExcelFormer
& 13.85 & - & 14.94 & 20.91
& 28.09 & - & 28.50 & 31.75
& 13.85 & - & 14.96 & 23.05 \\
\midrule
TransTab
& 19.83 & 19.09 & 22.34 & 21.04
& 43.86 & 45.27 & 34.57 & 39.99
& 55.44 & 55.86 & 55.75 & 55.93 \\
CARTE
& 25.80 & 30.30 & 22.17 & 27.27
& 54.42 & 55.82 & 54.99 & 51.92
& 55.91 & 57.13 & 55.60 & 55.53 \\
\midrule
TabICLv2
& 19.87 & - & 19.18 & 26.58
& 26.54 & - & 25.31 & 32.82
& 6.08 & - & 6.22 & 6.70 \\
TabPFNv3
& 20.56 & - & 20.22 & 21.86
& 30.85 & - & 31.48 & 34.57
& 4.29 & - & 5.05 & 6.26 \\
TabPFNv3-Plus
& 47.75 & - & 38.20 & 48.60
& 63.50 & - & 52.30 & 61.90
& 54.05 & - & 48.30 & 51.15 \\
\bottomrule
\end{tabular*}
}

\caption{Performance comparison on Join-based datasets (Accuracy, in \%).}
\label{tab:join_results}
\end{table*}

\subsection{Performance on Unionable Tables}
\subsubsection{\textbf{Overall Results}}
Table~\ref{tab:union_results} summarizes the experimental results under the union setting, from which we draw the following key observations.
First, regarding auxiliary table utilization strategies, the pre-training-based strategy (PT) yields the most substantial overall improvements, achieving a win rate of $66.7\%$ with an average gain of $0.34\%$\footnote{See Appendix~\ref{app:category_stats} for category-level Win Rate and Average Gain statistics.}.
This result indicates that, although target and auxiliary tables originate from different sources, they often correspond to the same task type and exhibit strong semantic relatedness in relevant fields, making them well suited for transfer learning approaches such as TransTab and CARTE.
Second, data augmentation (DA) and feature augmentation (FA) exhibit relatively limited benefits and may even degrade performance, with win rates of $33.4\%$ and $61.5\%$, and average gains of $-0.55\%$ and $0.27\%$, respectively.
This behavior is likely due to cross-source schema discrepancies and distribution shifts in this setting, which make reliable alignment and matching difficult, which introduces noise and leads to unstable gains.
Third, foundation models achieve the best average performance in the single-table setting, highlighting the strong transferability afforded by large-scale pre-training. 
However, they do not consistently benefit from any of the three auxiliary table utilization strategies, suggesting that enabling foundation models to effectively exploit auxiliary tables remains an important direction for future research.

\begin{figure}[t]
    \centering
    \begin{subfigure}[t]{0.49\columnwidth}
        \centering
        \includegraphics[width=\linewidth]{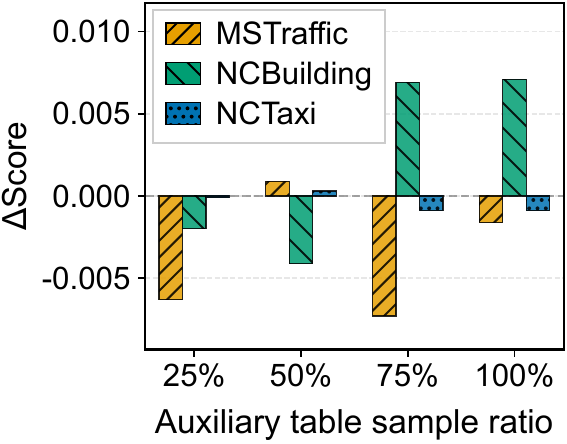}
        \caption{TransTab.}
        \label{fig:exp_union_pct_transtab}
    \end{subfigure}
    \hfill
    \begin{subfigure}[t]{0.49\columnwidth}
        \centering
        \includegraphics[width=\linewidth]{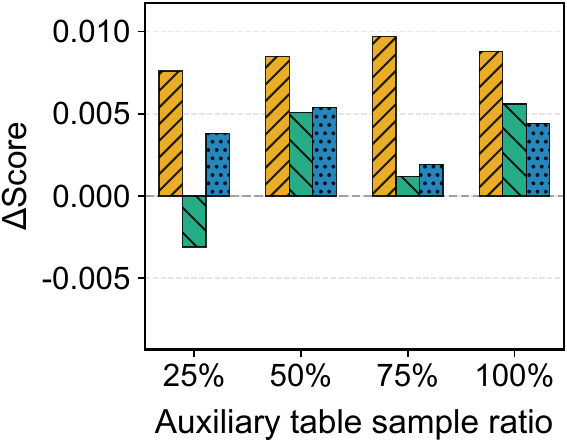}
        \caption{CARTE.}
        \label{fig:exp_union_pct_carte}
    \end{subfigure}

    \caption{Effect of auxiliary-table sample ratio on transfer performance (union setting). Bars report $\Delta$Score relative to the 0\% auxiliary-sample baseline (positive/negative values indicate gains/losses).}
    \label{fig:exp_union_pct}
\end{figure}

\subsubsection{\textbf{Effect of Auxiliary Data Volume on Pre-training}}
We further analyze how the proportion of auxiliary-table samples available during pre-training affects transfer performance in the union setting.
In the main experiments, all auxiliary samples are used for pre-training to maximize the potential for knowledge transfer.
To better characterize the relationship between auxiliary data volume and transfer gains, we partition the auxiliary tables into subsets, and use each subset for pre-training.
As shown in Figure~\ref{fig:exp_union_pct}, increasing the proportion of auxiliary samples generally leads to more stable positive gains or smaller negative deviations across most datasets for both TransTab and CARTE.
This trend indicates that the scale of auxiliary data is a key factor influencing transfer effectiveness in the union scenario.
However, mild negative gains or performance fluctuations are still observed at certain proportions, suggesting that when cross-source distribution shifts and weak semantic alignment coexist, increasing data volume alone does not guarantee monotonic improvements.
Developing methods to more effectively exploit large-scale auxiliary tables while mitigating transfer-induced noise represents a promising direction for future research.

\subsection{Performance on Joinable Tables}
\subsubsection{\textbf{Overall Results}}
Table~\ref{tab:join_results} reports the results under the join setting, from which we obtain the following observations.
First, regarding auxiliary table utilization strategies, the feature augmentation strategy (FA) delivers the most pronounced improvements, achieving a win rate of $87.2\%$ with an average gain of $2.78\%$.
This result suggests that feature augmentation is effective because rows aligned to the same real-world entity provide complementary rather than noisy information.
Second, transfer learning methods based on the pre-training strategy (PT) also demonstrate consistently strong performance, with a win rate of $83.3\%$ and an average improvement of $1.37\%$.
Third, foundation models with native support for textual columns, such as TabPFN v3-Plus, achieve the strongest overall performance. However, despite their success on single-table tasks, existing tabular foundation models generally fail to fully exploit information from auxiliary tables, indicating substantial room for improvement in extending foundation models to multi-table learning settings.

\subsubsection{\textbf{Impact of Matching Quality on Feature Augmentation}}
\begin{figure}[t]
\centering
\includegraphics[width=\linewidth]{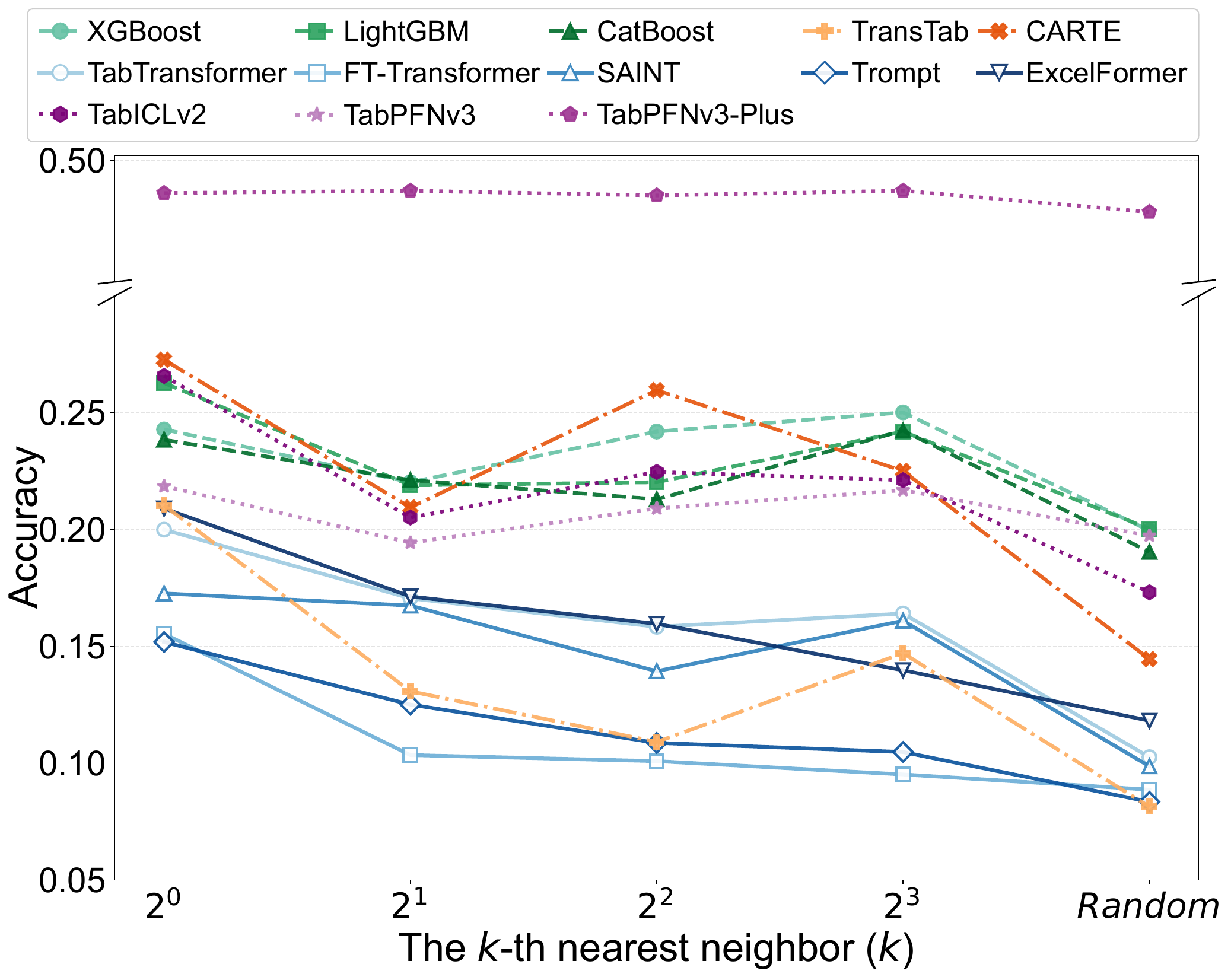} 
\caption{Impact of match quality on accuracy (join setting in NNStocks).}
\label{fig:exp_join_diff_k}
\end{figure}

We further investigate the impact of row-level matching quality on feature augmentation. Figure~\ref{fig:exp_join_diff_k} shows a clear performance degradation as matching quality decreases.
Although some fluctuations are observed, a clear overall trend emerges: as matching quality deteriorates, the amount of useful information contributed by the auxiliary table decreases, leading to a consistent decline in model performance.
This indicates that the usefulness of auxiliary features diminishes when row alignment becomes less accurate, leading to lower classification accuracy.
Overall, these results show that feature augmentation in the join setting critically relies on high-quality one-to-one row matching, highlighting row-level matching as an important direction for future join-based data lake learning.

\section{Related Work}
\label{sec:related_work}


\subsection{Single-Table Tabular Benchmarks}
Traditional tabular benchmarks assume a single, self-contained table in which each row represents an independent instance. Early studies primarily relied on datasets from the UCI Machine Learning Repository~\cite{asuncion2007uci} and Kaggle competitions, but these often lacked standardized preprocessing and evaluation protocols.
OpenML addressed this limitation by providing large-scale benchmark datasets with unified task definitions, standardized train--test splits, and rich metadata~\cite{bischl2openml}. More recent benchmarks further improve realism from different perspectives. TabArena introduces a continuously evolving benchmarking platform with dynamic datasets and leaderboards~\cite{erickson2025tabarena}. TabReD incorporates temporal dynamics and richer domain characteristics~\cite{rubachev2025tabred}, while TableShift~\cite{gardner2023tableshift} and TabFSBench~\cite{cheng2025tabfsbench} evaluate model robustness under realistic feature distribution shifts.
Despite these advances, existing benchmarks assume that training and inference are performed on a single, pre-integrated table, overlooking the challenges of integrating heterogeneous data sources before model learning.

\subsection{Multiple-Table Tabular Benchmarks}

Recent work has explored multi-table learning from two perspectives: schema-agnostic cross-table generalization and schema-aware relational learning.
The first line of work learns transferable representations across heterogeneous tables without explicitly modeling their relationships. TransTab~\cite{wang2022transtab} aligns column semantics in a shared embedding space, XTab~\cite{zhu2023xtab} learns transferable representations via large-scale contrastive pre-training, and CARTE~\cite{kim2024carte} captures cross-table context using heterogeneous graph attention networks.
The second line explicitly exploits relational schemas. RelBench~\cite{robinson2024relbench} provides standardized relational learning tasks over multi-table databases, while 4DBInfer~\cite{wang2024_4dbinfer} focuses on graph-based reasoning in relational environments. SJTUTables extends classical datasets such as MovieLens and LastFM into multi-table classification benchmarks~\cite{li2024rllm}.
Although these benchmarks demonstrate the benefits of relational modeling, they assume clean and curated relational databases with explicit schema relationships.

\subsection{Discussion}

Existing benchmarks provide comprehensive evaluations for single-table learning and relational databases. However, benchmarks for model learning over data lakes remain largely unexplored, particularly under unionable and joinable multi-table settings. LakeMLB addresses this gap by modeling heterogeneous table collections, flexible multi-table integration through union and join operations, and diverse downstream learning tasks.
\section{Conclusion and Future Work}
\label{sec:conclusion}
We introduce LakeMLB, the first benchmark for evaluating machine learning over multi-table data in data lake environments. Covering both Union and Join scenarios across diverse real-world datasets, LakeMLB captures the key challenges of multi-table learning in data lakes. Extensive experiments show that data integration strategies are highly scenario-dependent: pre-training performs best for Union tasks, feature augmentation excels on Join tasks, and transfer learning demonstrates strong overall robustness. We hope LakeMLB will serve as a standard benchmark and stimulate the development of scalable and robust learning methods for data lake and lakehouse systems. Future work will extend LakeMLB to more realistic settings with larger and more heterogeneous table collections, richer relational structures, evolving schemas, and broader downstream learning tasks.

\bibliography{references}

\clearpage
\appendix

\section{Benchmark Design Details}
\label{app:benchmark}

\subsection{Unionable Tables Details}
The unionable setting comprises three datasets spanning three distinct application domains. In each dataset, the target and auxiliary tables describe the same underlying prediction scenario and exhibit closely related, though not necessarily identical, label semantics, making them suitable for \textit{union-style} integration.

\noindent \subsubsection*{\textbf{MSTraffic} (\textbf{M}aryland and \textbf{S}eattle \textbf{Traffic} Collisions)}
Both tables are collected from the United States Government's open data portal\footnote{\url{https://data.gov/}} and describe urban traffic collision events.
\textbf{Target Table (\textit{Maryland}):} The target table, denoted as \textit{Maryland}, is derived from Maryland's crash reporting incidents data\footnote{\url{https://catalog.data.gov/dataset/crash-reporting-incidents-data}}. We select records from January 2017 to December 2023, resulting in 10,800 samples with 37 features capturing environmental and contextual conditions at the time of the crash (e.g., weather, roadway condition, and lighting). The task is to classify \textit{Collision Type} into 9 categories.
\textbf{Auxiliary Table (\textit{Seattle}):} The auxiliary table, denoted as \textit{Seattle}, is obtained from the Seattle Department of Transportation collision records\footnote{\url{https://catalog.data.gov/dataset/sdot-collisions-all-years-2a008}}. It contains 10,800 samples with 50 features, providing richer metadata such as collision severity, casualty counts, risky driving behaviors, and road structure attributes. This table also includes a collision-type label column, \textit{COLLISIONTYPE}, with 9 categories. While the label space partially overlaps with that of \textit{Maryland}, the two annotation systems exhibit minor semantic discrepancies due to differences in local reporting standards.

\noindent \subsubsection*{\textbf{NCBuilding} (\textbf{N}ew York and \textbf{C}hicago \textbf{Building} Violations)}
Both tables are sourced from the United States Government's open data portal and focus on building inspection and violation reporting in metropolitan areas.
\textbf{Target Table (\textit{NewYork}):} The target table, denoted as \textit{NewYork}, is based on New York City housing maintenance code violations\footnote{\url{https://catalog.data.gov/dataset/housing-maintenance-code-violations}}. We retain records from January 2023 to December 2024 and obtain 30,000 samples after filtering, with 40 features including registration metadata and administrative location attributes. The task is to predict \textit{Violation Category} among 30 classes.
\textbf{Auxiliary Table (\textit{Chicago}):} The auxiliary table, denoted as \textit{Chicago}, corresponds to Chicago building violation records\footnote{\url{https://catalog.data.gov/dataset/building-violations}} over the same time span (2023--2024). It contains 37,000 samples with 23 features (e.g., violation descriptions and inspector-related information) and covers 37 violation categories. The label taxonomy differs from that of \textit{NewYork} due to city-specific regulatory systems and reporting practices.

\noindent \subsubsection*{\textbf{NCTaxi} (\textbf{N}ew York and \textbf{C}hicago \textbf{Taxi} Trips)}
Both tables are sourced from NYC Open Data and the Chicago Data Portal and describe urban taxi trip records.
\textbf{Target Table (\textit{NewYork}):} The target table, denoted as \textit{NewYork}, is derived from the New York City Yellow Taxi Trip Data\footnote{\url{https://data.cityofnewyork.us/Transportation/2023-Yellow-Taxi-Trip-Data/4b4i-vvec}}. We select records from January to December 2023 and obtain 100,000 samples after filtering, with 19 features describing trip times, passenger counts, travel distances, pickup and drop-off Taxi Zones, payment methods, fares, tips, taxes, tolls, and surcharges. The task is to predict the drop-off Taxi Zone, denoted as \textit{dolocationid}, among 50 classes.
\textbf{Auxiliary Table (\textit{Chicago}):} The auxiliary table, denoted as \textit{Chicago}, is obtained from the Chicago Taxi Trips data portal\footnote{\url{https://data.cityofchicago.org/Transportation/Taxi-Trips-2023/e55j-2ewb}}. We select records from January to December 2023, resulting in 100,000 samples with 19 features, including trip duration, travel distance, pickup community area, fare components, payment method, taxi company, and pickup location information. Four drop-off-specific geographic attributes are removed to prevent label leakage. This table includes a drop-off community-area label, \textit{dropoff\_community\_area}, covering 50 classes. Although both label spaces represent trip destinations, they follow different city-specific geographic zoning systems and therefore exhibit some semantic differences.

\subsection{Joinable Tables Details}
The joinable setting likewise comprises three datasets spanning three distinct application domains. In each dataset, the auxiliary table provides complementary attributes for enriching the target table through \textit{join-style} integration. When explicit primary--foreign-key relationships are unavailable, we establish weak joins using fuzzy entity matching.

\noindent \subsubsection*{\textbf{NNStocks} (\textbf{N}ASDAQ and \textbf{N}YSE \textbf{Stocks})}
This dataset focuses on sector classification for publicly listed U.S. companies.
\textbf{Target Table (\textit{NNList}):} The target table, denoted as \textit{NNList}, is constructed from the official NASDAQ and NYSE listings (December 2025)\footnote{\url{https://github.com/rreichel3/US-Stock-Symbols}}. We retain companies with available Wikipedia pages, resulting in 1,078 samples with 11 features. The task is to predict \textit{Sector Classification} over 11 categories.
\textbf{Auxiliary Table (\textit{NNWiki}):} The auxiliary table, denoted as \textit{NNWiki}, is collected by extracting corporate infobox attributes from Wikipedia\footnote{\url{https://www.wikipedia.org/}}. It contains 937 samples; after removing columns with sparsity greater than 95\%, we obtain 22 informative features. Since there is no explicit foreign key between \textit{NNList} and \textit{NNWiki}, we construct a weak join via fuzzy entity matching: we encode the \textit{name} field in \textit{NNList} and the \textit{wiki title} field in \textit{NNWiki} using BERT-base-uncased, and perform 1-nearest-neighbor (1-NN) retrieval in the embedding space with cosine similarity as the matching criterion. This yields an entity mapping along with similarity scores.

\noindent \subsubsection*{\textbf{DSMusic} (\textbf{D}iscogs and \textbf{S}potify \textbf{Music})}
This dataset targets music genre classification with cross-platform metadata integration.
\textbf{Target Table (\textit{Discogs}):} The target table, denoted as \textit{Discogs}, is sourced from a public Discogs database snapshot\footnote{\url{https://www.kaggle.com/datasets/fleshmetal/records-a-comprehensive-music-metadata-dataset}}. After preprocessing, which includes removing the \textit{styles} column to avoid potential label leakage and performing balanced sampling over 11 single-label classes, the table contains 11,000 samples with 5 features. The task is to predict \textit{Genre Classification} over 11 categories.
\textbf{Auxiliary Table (\textit{Spotify}):} The auxiliary table, denoted as \textit{Spotify}, is obtained from a Spotify tracks dataset\footnote{\url{https://www.kaggle.com/datasets/maharshipandya/-spotify-tracks-dataset}}. It contains 11,000 samples with 21 features that provide richer audio and popularity metadata (e.g., track popularity, rhythm structure, and duration). To enable joining, we encode the \textit{title} field in \textit{Discogs} and the \textit{track name} field in \textit{Spotify} using BERT-base-uncased, and apply cosine-similarity-based 1-NN retrieval to obtain record-level alignments and similarity scores.

\noindent \subsubsection*{\textbf{AGBooks} (\textbf{A}mazon and \textbf{G}oodreads \textbf{Books})}
This dataset targets book category classification through cross-platform metadata integration.
\textbf{Target Table (\textit{Amazon}):} The target table, denoted as \textit{Amazon}, is constructed from the book metadata in the Amazon Reviews 2023 collection released by the McAuley Lab\footnote{\url{https://mcauleylab.ucsd.edu/public_datasets/data/amazon_2023/}}. We select 40 book categories and perform balanced sampling of 2,500 records per category, resulting in 100,000 samples with 11 features. The table provides product-level metadata, including product identifiers, titles, average ratings, rating counts, prices, author or store information, descriptions, and product attributes. The task is to predict \textit{Book Category Classification} over 40 categories.
\textbf{Auxiliary Table (\textit{Goodreads}):} The auxiliary table, denoted as \textit{Goodreads}, is obtained from the Goodreads book datasets released by the McAuley Lab\footnote{\url{https://cseweb.ucsd.edu/~jmcauley/datasets/goodreads.html}}. It contains 100,000 samples with 20 features that provide complementary bibliographic and community metadata, including average ratings and review counts, publication dates, publishers, page counts, languages, formats, ISBN identifiers, ebook indicators, author identifiers, similar books, and descriptions. Since there is no complete and reliable foreign key between \textit{Amazon} and \textit{Goodreads}, we construct a weak join via fuzzy entity matching: we encode the \textit{title} fields in both tables using BERT-base-uncased, and perform 1-nearest-neighbor (1-NN) retrieval in the embedding space with cosine similarity as the matching criterion. This yields record-level Amazon--Goodreads mappings along with similarity scores.

\section{Task Challenges and Potential Solutions}
\label{app:chall_sol}

In this section, we discuss the key challenges of the benchmark tasks and potential solutions.

\subsection{Task Challenges}

Machine learning over \emph{unionable} and \emph{joinable} tables presents a set of fundamental challenges that stem from the intrinsic characteristics of large-scale, multi-source tabular data. We summarize the key challenges as follows.

\begin{enumerate}
    \item \textbf{Data heterogeneity.}
    Tables may differ substantially in schemas, data types, value formats, and underlying statistical distributions. This heterogeneity is further amplified in multi-source environments, making cross-table alignment and unified modeling particularly challenging.

    \item \textbf{Data quality.}
    Real-world tabular data usually originates from diverse sources, ranging from carefully curated enterprise databases to collaboratively constructed web tables. The lack of consistent design principles and quality control mechanisms leads to pervasive noise, inconsistencies, redundancy, and missing values, all of which can severely impair training stability and model generalization.

    \item \textbf{Semantic gaps.}
    Tables are often created for different application domains, resulting in substantial discrepancies in data organization, semantic interpretation, and implicit assumptions. Such domain-level semantic gaps hinder effective knowledge sharing and transfer across tables.

    \item \textbf{Scalability limitations.}
    Data lakes are characterized by massive scale and high dimensionality. Joint learning over large collections of unionable or joinable tables requires machine learning algorithms that are both computationally efficient and highly scalable, which remains challenging in practice.
\end{enumerate}

\subsection{Potential Solutions}
\label{subsec:pot_solutions}

Most existing tabular learning methods are not explicitly designed for learning over \emph{unionable} or \emph{joinable} tables. Under simplified assumptions or relaxed formulations, however, they can still serve as reasonable baselines.
A naive alternative is to ignore auxiliary tables and directly apply standard tabular learning models to the target table.
In the following, we discuss how auxiliary tables can be leveraged to improve performance on the target task.

\subsubsection{\textbf{Learning with Unionable Tables}}

As illustrated in Figure~\ref{fig_data_example}, unionable tables can be viewed as providing additional training instances, although their feature and label spaces may differ substantially from those of the target table. Under this setting, two representative classes of approaches can be considered.

\textbf{(1) Pre-training-based strategies} aim to transfer knowledge learned from auxiliary domains to the target task, typically through shared representations or model parameters. In this context, a model can be pre-trained on auxiliary tables, either in a supervised or unsupervised manner, and then fine-tuned on the target table.
However, this paradigm faces several challenges. Feature spaces across tables are often inconsistent, violating assumptions made by many transfer learning methods. Moreover, defining appropriate pre-training objectives is non-trivial: supervised pre-training requires identifying label columns with consistent semantics, while unsupervised objectives are complicated by strong intra-table and inter-table heterogeneity.

\textbf{(2) Data augmentation-based strategies} expand the training set by assigning pseudo labels to unlabeled samples and jointly training on both labeled and pseudo-labeled data. In the unionable setting, samples with similar semantics can be retrieved from auxiliary tables and treated as pseudo-labeled instances. The main challenge lies in pseudo-label quality, as label semantics often vary across tables, making accurate alignment difficult and introducing potential noise.

\subsubsection{\textbf{Learning with Joinable Tables}}
As shown in Figure~\ref{fig_data_example}, joinable tables provide additional feature dimensions for each instance, naturally leading to a \textbf{feature augmentation} paradigm.
In practice, this can be implemented using a simple 1-NN-based feature augmentation approach. Specifically, for each row in the target table, a semantic primary key corresponding to a real-world entity is identified. Matching rows from auxiliary tables are then retrieved based on this key and concatenated with the target row along the feature dimension. Standard tabular learning models can be directly applied to the resulting augmented table. The key challenge is accurate row matching, as incorrect joins introduce noisy features that can significantly degrade performance.


\section{Auxiliary-table Strategies (PT/DA/FA)}
\label{app:aux_strategies}

This appendix systematically describes three strategies for leveraging auxiliary tables to improve predictive performance on the target table: pre-training-based strategy (PT), data augmentation-based strategy (DA), and feature augmentation-based strategy (FA). To ensure interpretability and reproducibility in controlled comparisons, we adopt designs with low implementation complexity and well-isolated variables. Unless otherwise specified, all methods are evaluated exclusively on the fixed test set of the target table; auxiliary tables are used only for pre-training, training-time sample augmentation, or feature supplementation, and never enter the test or evaluation stage.

\subsubsection*{\textbf{Basic Setting}}
\label{app:aux_strategies_basic_setting}
To enable fair comparisons across strategies, we construct a fixed train/validation/test split for the target table of each dataset and reuse the same split across all methods. All preprocessing operations (e.g., normalization/standardization for numerical features, vocabulary construction for categorical encoding, and missing value handling) are determined solely from the target-table training set. All strategies are compared under the same model architecture and training budget, with the number of training epochs, optimizer configuration, and early-stopping criterion kept identical, so that performance differences can be primarily attributed to the way auxiliary-table information is utilized.

\subsection{Pre-training-based Strategy}
\label{app:aux_strategies_pt}

PT is applicable to all six datasets in this benchmark because it does not require the target and auxiliary tables to share an identical label space. The key idea is to first pre-train on the auxiliary table to obtain a more generalizable parameter initialization and representation, and then fine-tune the model on the target table under the supervised objective, thereby improving downstream classification performance on the target-table test set.

PT follows a two-stage pipeline, \emph{auxiliary-table pre-training} $\rightarrow$ \emph{target-table fine-tuning}, as summarized below:
\begin{itemize}
    \item \textbf{Auxiliary-table pre-training:} We pre-train the model on the full auxiliary table. The pre-training objective and training details follow the default configuration of the corresponding method (e.g., self-supervised representation learning for tabular data).
    \item \textbf{Target-table fine-tuning:} We fine-tune the model on the target-table training set, and use the target-table validation set for early stopping and hyper-parameter selection. Final results are reported only on the target-table test set.
\end{itemize}

\subsection{Data Augmentation-based Strategy}
\label{app:aux_strategies_da}

DA improves supervised learning on the target table by introducing additional training samples. To keep the strategy simple, reproducible, and easy to attribute, DA consists of three steps:

\begin{itemize}
    \item \textbf{Coarse schema alignment:}
    We first perform column-level coarse alignment by normalizing column names (matching ignores differences in letter case and underscores). Columns with matched names are mapped to the same fields; the remaining auxiliary columns are retained as additional fields, with the corresponding entries in the target table filled as missing values, resulting in a unified schema that can be processed by a single pipeline. For numerical features, normalization/ standardization statistics are estimated only from the target-table training set. For categorical features, the encoding vocabulary is also built from the target-table training set, and unseen values are mapped to a unified unknown token. Missing values are represented with a consistent placeholder to ensure that samples from both tables can be ingested uniformly.

    \item \textbf{Label alignment:}
    \begin{itemize}
        \item \emph{Unionable setting: label-space mapping.}
        We use the label set of the target table as the reference. Category names from both tables are encoded using BERT-base-uncased, and label alignment is performed via 1-nearest-neighbor (1-NN) retrieval in the embedding space with cosine similarity. The auxiliary samples are then relabeled into the target-table taxonomy. This ensures that the supervision signal introduced by DA is consistent with the target task definition and facilitates attribution of performance gains.

        \item \emph{Joinable setting: pseudo-label construction.}
        Since the auxiliary table does not contain an explicit label column, we first establish cross-table 1-NN pairing based on sample identifiers, and then assign the target-table labels to the matched auxiliary samples. The matched auxiliary samples are deduplicated and used for training-time sample expansion.
    \end{itemize}
    \item \textbf{Sample appending:}
    After schema and label alignment, we append the retained auxiliary samples row-wise to the target-table training set to form an augmented training set. To control augmentation strength and maintain comparability across datasets, the number of appended samples is capped at 30\% of the target-table training size. The validation and test sets remain unchanged. We then train the model on the augmented training set and report results only on the target-table test set.
\end{itemize}

To avoid biased comparisons caused by inconsistent augmentation strength, DA adopts a unified sample-appending cap (30\%) across all datasets and always aligns supervision signals to the target-table taxonomy. Consequently, the gains of DA primarily stem from effective training-time sample expansion rather than uncontrolled changes in evaluation protocol.

\subsection{Feature Augmentation-based Strategy}
\label{app:aux_strategies_fa}

FA is applicable to all six datasets in this benchmark. Unlike DA, which increases the number of training samples, FA augments the feature space without changing the number of target-table rows. Specifically, we construct a record-level weak linkage such that each target-table sample is associated with its most similar candidate in the auxiliary table, and then concatenate the auxiliary attributes to form an augmented feature representation. For datasets with natural entity anchors, we perform entity-name matching; otherwise, we adopt row-text matching.

\begin{itemize}
    \item{Joinable: entity-name 1NN matching.}
    For \textit{NNStocks}, \textit{DSMusic}, and \textit{AGBooks}, we use entity names as the anchor for weak linkage. We encode the target-table entity-name columns (e.g., \textit{name}, \textit{Company Name}, \textit{title}) and the auxiliary-table entity-name columns (e.g., \textit{wiki title}, \textit{track name}) using BERT-base-uncased, and retrieve the auxiliary 1-NN for each target record in the embedding space. We then concatenate only the auxiliary attribute features to the corresponding target rows to achieve feature augmentation. Finally, we train a single-table model on the augmented target table and evaluate on the target-table test set.

    \item{Unionable: row-text 1NN matching.}
    For \textit{MSTraffic}, \textit{NCBuilding}, and \textit{NCTaxi}, explicit keys suitable for direct joins are unavailable. We therefore apply record-level weak matching to inject auxiliary features. Concretely, we represent each row as a textual sequence by concatenating \emph{column name--cell value} pairs, encode these row texts, and perform 1-NN matching in the embedding space to select the most similar auxiliary candidate for each target record. We then concatenate the auxiliary candidate's attribute features to the target record, thereby expanding the target-table feature space.
\end{itemize}

\paragraph{Semantic Representation-Based 1-NN Matching.}
The two weak-link construction schemes described above follow a unified semantic nearest-neighbor matching procedure. First, the fields or records used for matching in the target and auxiliary tables are converted into textual sequences and encoded using the pre-trained BERT-base-uncased model. We extract the hidden representation of the \texttt{[CLS]} token from the final layer as the holistic semantic representation of each record. The embeddings of both tables are then $L_2$-normalized, and a FAISS HNSW approximate nearest-neighbor index is constructed over the auxiliary-table embeddings. For each target record, its semantic embedding is used as a query to perform 1-NN retrieval over the auxiliary table, and the auxiliary record with the highest semantic similarity is selected as its match. The attributes of the matched auxiliary record are subsequently concatenated with those of the corresponding target record to produce the feature-augmented target table. Because Euclidean distance and cosine similarity induce the same ranking over $L_2$-normalized vectors, the nearest neighbor returned by HNSW can be directly used for semantic matching.

\paragraph{Implementation Details.}
We load BERT-base-uncased from a local checkpoint and truncate or pad
each input sequence to a maximum length of 128 tokens. The FAISS HNSW index uses
$M=32$, $\textit{efConstruction}=64$, and $\textit{efSearch}=64$.
The random-matching baseline uses a fixed seed of 42. No explicit random
seed is used in the nearest-neighbor ranking step.

Regarding computational complexity, let $n_1$ and $n_2$ denote the numbers of records in the target and auxiliary tables, respectively. When the BERT architecture, maximum input length, and embedding dimensionality are fixed, the costs of text encoding and vector normalization grow linearly with the total number of records. After semantic encoding, the $n_2$ auxiliary embeddings are incrementally inserted into the HNSW graph index. Under the standard average-case analysis of HNSW, the graph traversal required for each insertion grows approximately logarithmically with the auxiliary-table size, resulting in an index-construction cost of approximately $n_2\log n_2$. During retrieval, a single nearest-neighbor query over an index containing $n_2$ candidates incurs an average cost of approximately $\log n_2$. Performing this search for all $n_1$ target records therefore incurs a query cost of approximately $n_1\log n_2$. Accordingly, the overall average complexity of index construction and nearest-neighbor retrieval is $O((n_1+n_2)\log n_2)$. When the two tables are of comparable size, such that both $n_1$ and $n_2$ can be represented by $n$, the average complexity of the matching procedure can be further approximated as $O(n\log n)$. This complexity is lower than the quadratic complexity required by exhaustive pairwise comparison, thereby reducing the computational cost of large-scale table matching. We note that this analysis characterizes the typical average-case behavior of HNSW rather than providing a strict worst-case complexity guarantee.

FA uniformly uses 1-NN as the weak-link construction rule and performs matching once before training to generate the augmented feature table. During training and evaluation, the number of target-table samples remains unchanged, ensuring that performance differences primarily reflect feature augmentation rather than changes in sample size.

\section{Experimental Details}
\label{app:exp_details}

\subsection{Compute Environment}
All experiments are conducted on a server running CentOS Linux 7 (Core) with Linux kernel 3.10.0-1160.90.1.el7.x86\_64.
The server is equipped with an Intel Xeon E5-2630 v4 @ 2.20GHz CPU, 188 GB RAM, and 2 $\times$ NVIDIA RTX A6000 GPUs (49{,}140 MiB $\approx$ 48 GB per GPU), using NVIDIA driver 550.67 and CUDA 12.4. The software environment consists of Python 3.9.21, PyTorch 2.8.0,
scikit-learn 1.6.1, XGBoost 2.1.4, LightGBM 4.6.0,
CatBoost 1.2.8, Transformers 4.57.3, FAISS-CPU 1.13.0,
TabPFN 8.0.7 (TabPFN V3), TabICL 2.0.2 (TabICLv2), and
PyTorch Geometric 2.6.1.

\subsection{Data Splits and Leakage Prevention}
Each target table is partitioned using a fixed mask into approximately
70\% training, 10\% validation, and 20\% test instances. The same split masks are used across all methods and are publicly available in our open-source repository to support reproducible evaluation. All data-dependent preprocessing operations, including
numerical normalization, categorical vocabulary construction, and
missing-value handling, are fitted exclusively on the training split
and subsequently applied unchanged to the validation and test splits.
Hyperparameters are selected solely according to validation-set
performance, and the test split is not evaluated during model or
hyperparameter selection.

For TabPFN and TabICL, after fixing the model configuration, we use the
union of the training and validation splits as the final fitting context
for test-set prediction; test labels are never accessed. For the DA
strategy, target labels used for label alignment and augmented-sample
construction are drawn exclusively from the training split, while the
validation and test splits remain unchanged.

\subsection{Hyper-parameter Selection Protocol}
We select hyper-parameters based on validation-set performance.
To ensure fair comparisons while mitigating excessive validation overfitting, we adopt a unified grid-search protocol for tree-based and deep tabular baselines.
For transfer learning and tabular foundation model baselines, we follow the official default configurations and only tune a limited set of task-sensitive hyper-parameters when applicable to avoid disproportionate tuning advantages, since extensive search is computationally and memory intensive and typically yields limited marginal gains.
Unless otherwise specified, iterative learners use early stopping.
We evaluate each locally executed model--dataset configuration over ten
independent runs with independently generated random seeds.
Hyperparameter search uses a fixed seed of 42. TabPFNv3-Plus is evaluated
through its official online platform. For every reported TabPFNv3-Plus
result, we perform ten separate platform executions and report their mean.
Because the platform does not expose random-seed control, these executions
use the platform's default settings and are excluded from the controlled-seed
protocol.

\begin{table*}[!t]
  \centering
  \setlength{\tabcolsep}{10pt}

  \begin{tabular}{l l c c c c c c}
    \toprule
    \multirow{2}{*}{\textbf{Relation}}
    & \multirow{2}{*}{\textbf{Category}}
    & \multicolumn{2}{c}{\textbf{+PT($T_{aux}$)}}
    & \multicolumn{2}{c}{\textbf{+DA($T_{aux}$)}}
    & \multicolumn{2}{c}{\textbf{+FA($T_{aux}$)}} \\
    \cmidrule(lr){3-4}
    \cmidrule(lr){5-6}
    \cmidrule(lr){7-8}

    & & \textbf{Win\%} & \textbf{Avg. Gain}
      & \textbf{Win\%} & \textbf{Avg. Gain}
      & \textbf{Win\%} & \textbf{Avg. Gain} \\
    \midrule
    \multirow{5}{*}{\textbf{Union}}
    & Classical Tree-based & -- & -- & 66.67 & 0.37 & 66.67 & 0.87 \\
    & Single-table TNNs    & -- & -- & 40.00 & -0.61 & 66.67 & 0.45 \\
    & Transfer Learning    & 66.67 & 0.34 & 0.00 & -1.02 & 33.34 & -0.28 \\
    & Foundation Models    & -- & -- & 11.11 & -1.05 & 66.67 & -0.25 \\
    \cmidrule(lr){2-8}
    & \textit{Overall}     & \textbf{66.67} & \textbf{0.34}
                              & 33.34 & -0.55
                              & 61.54 & 0.27 \\

    \midrule

    \multirow{5}{*}{\textbf{Join}}
    & Classical Tree-based & -- & -- & 33.34 & -0.16 & 100.00 & 2.17 \\
    & Single-table TNNs    & -- & -- & 46.67 & -0.62 & 100.00 & 5.04 \\
    & Transfer Learning    & 83.33 & 1.37 & 50.00 & -1.64 & 50.00 & -0.60 \\
    & Foundation Models    & -- & -- & 33.34 & -3.03 & 77.78 & 1.88 \\
    \cmidrule(lr){2-8}
    & \textit{Overall}     & 83.33 & 1.37
                              & 41.03 & -1.23
                              & \textbf{87.18} & \textbf{2.78} \\

    \bottomrule
  \end{tabular}

  \caption{Performance comparison of method categories across the
  Union and Join settings. We report the Win Rate (Win\%) and
  Average Gain (Avg. Gain, in percentage points) for the PT, DA,
  and FA strategies to measure the consistency and magnitude of
  performance improvements.}
  \label{tab:main_results}
\end{table*}

\subsection{Hyper-parameter Search Space}
\vspace{2pt}
\noindent\textbf{Tree-based baselines.}
We tune tree-based models via grid search on the validation set.
All tree models use early stopping with patience set to 50 validation rounds.
The search spaces are:
\begin{itemize}
    \item \textbf{XGBoost:} n\_estimators $\in \{300, 500, 1000\}$, max\_depth $\in \{4, 6, 8\}$, learning\_rate $\in \{0.01, 0.03, 0.05\}$, subsample $\in \{0.8, 0.9\}$, colsample\_bytree $\in \{0.8, 0.9\}$.
    \item \textbf{CatBoost:} iterations $\in \{300, 500, 1000\}$, depth $\in \{4, 6, 8\}$, learning\_rate $\in \{0.01, 0.03, 0.05\}$.
    \item \textbf{LightGBM:} num\_boost\_round $\in \{300, 500, 1000\}$, num\_leaves $\in \{31, 63, 127\}$, learning\_rate $\in \{0.01, 0.03, 0.05\}$, feature\_fraction $\in \{0.8, 0.9\}$, bagging\_fraction $\in \{0.8, 0.9\}$.
\end{itemize}

\vspace{2pt}
\noindent\textbf{Deep tabular baselines.}
We tune deep tabular models via grid search on the validation set with shared training settings:
batch size fixed to 256, maximum epochs fixed to 500, and early stopping with patience set to 10.
The shared search space is:
\begin{itemize}
    \item hidden dimension $\in \{32, 64, 128\}$;
    \item number of layers $\in \{2, 3, 4\}$;
    \item learning rate $\in \{10^{-3}, 5\times 10^{-4}, 10^{-4}\}$;
    \item weight decay $\in \{10^{-4}, 5\times 10^{-4}, 10^{-3}\}$.
\end{itemize}

Method-specific settings are as follows:
\begin{itemize}
    \item \textbf{FT-Transformer:} no additional hyper-parameters are tuned beyond the shared search space.
    \item \textbf{TabTransformer:} the number of attention heads is fixed to num\_heads $= 8$.
    \item \textbf{SAINT:} num\_feats is determined by dataset metadata (i.e., total number of columns) rather than tuned.
    \item \textbf{Trompt:} the prompt pool size is fixed to num\_prompts $= 128$.
    \item \textbf{ExcelFormer:} no additional hyper-parameters are tuned beyond the shared search space.
\end{itemize}

For each evaluated configuration, we compute the mean and standard
deviation across the ten runs to quantify average performance and
run-to-run variability.

\vspace{2pt}
\noindent\textbf{Transfer learning-based baselines.}
For transfer learning-based methods, we follow the default configurations from the original implementations and only tune the most task-sensitive hyper-parameters on the validation set:
\begin{itemize}
    \item \textbf{TransTab:} pretrain\_epoch $\in \{20, 40, 60, 80, 100\}$.
    \item \textbf{CARTE:} target\_fraction $\in \{0.125, 0.25, 0.5\}$.
\end{itemize}
All other hyper-parameters follow the default settings of each method.

\vspace{2pt}
\noindent\textbf{Tabular foundation model baselines.}
For tabular foundation model baselines, we follow the default configurations from the official implementations and do not perform grid search.
\begin{itemize}
    \item \textbf{TabICLv2:} We use the default TabICLv2 checkpoint with
    $n_{\mathrm{estimators}}=8$ and $\mathrm{batch\_size}=8$. We enable
    many-class support, and retain the default softmax
    temperature of $0.9$ and outlier threshold of $4.0$.
    \item \textbf{TabPFNv3:} We use the default TabPFN V3 classifier with
    $n_{\mathrm{estimators}}=8$, $\mathrm{fit\_mode}=\mathrm{fit\_preprocessors}$,
    and built-in preprocessing. We set
    $\mathrm{ignore\_pretraining\_limits}=\mathrm{True}$ to evaluate the full
    input tables; all other inference settings follow the defaults.
    \item \textbf{TabPFNv3-Plus:} Due to limitations on API access, we evaluate TabPFNv3-Plus through its official online playground\footnote{\url{https://ux.priorlabs.ai/playground}}. Owing to the input-size constraints of the web interface, datasets containing more than 40,000 samples are reduced to 10,000 samples through proportionate stratified random subsampling, where the same sampling ratio is applied to each class to preserve the original class distribution. All hyperparameters are set to the default values recommended by the official platform.
\end{itemize}

\begin{figure}[t]
\centering
\includegraphics[width=\linewidth]{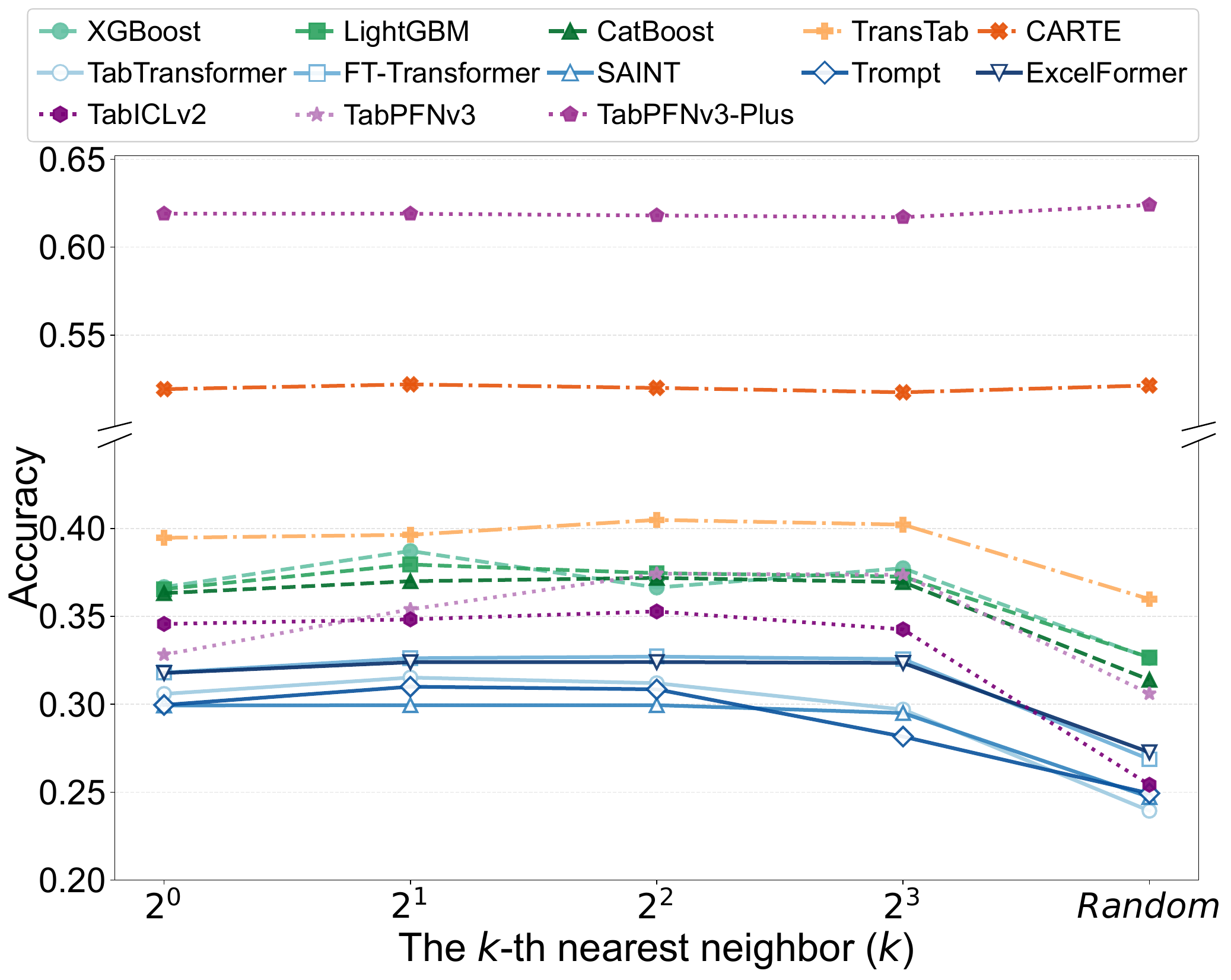} 
\caption{Impact of match quality on accuracy (join setting in DSMusic).}
\label{fig:exp_join_diff_k_dsmusic}
\end{figure}

\begin{figure}[t]
\centering
\includegraphics[width=\linewidth]{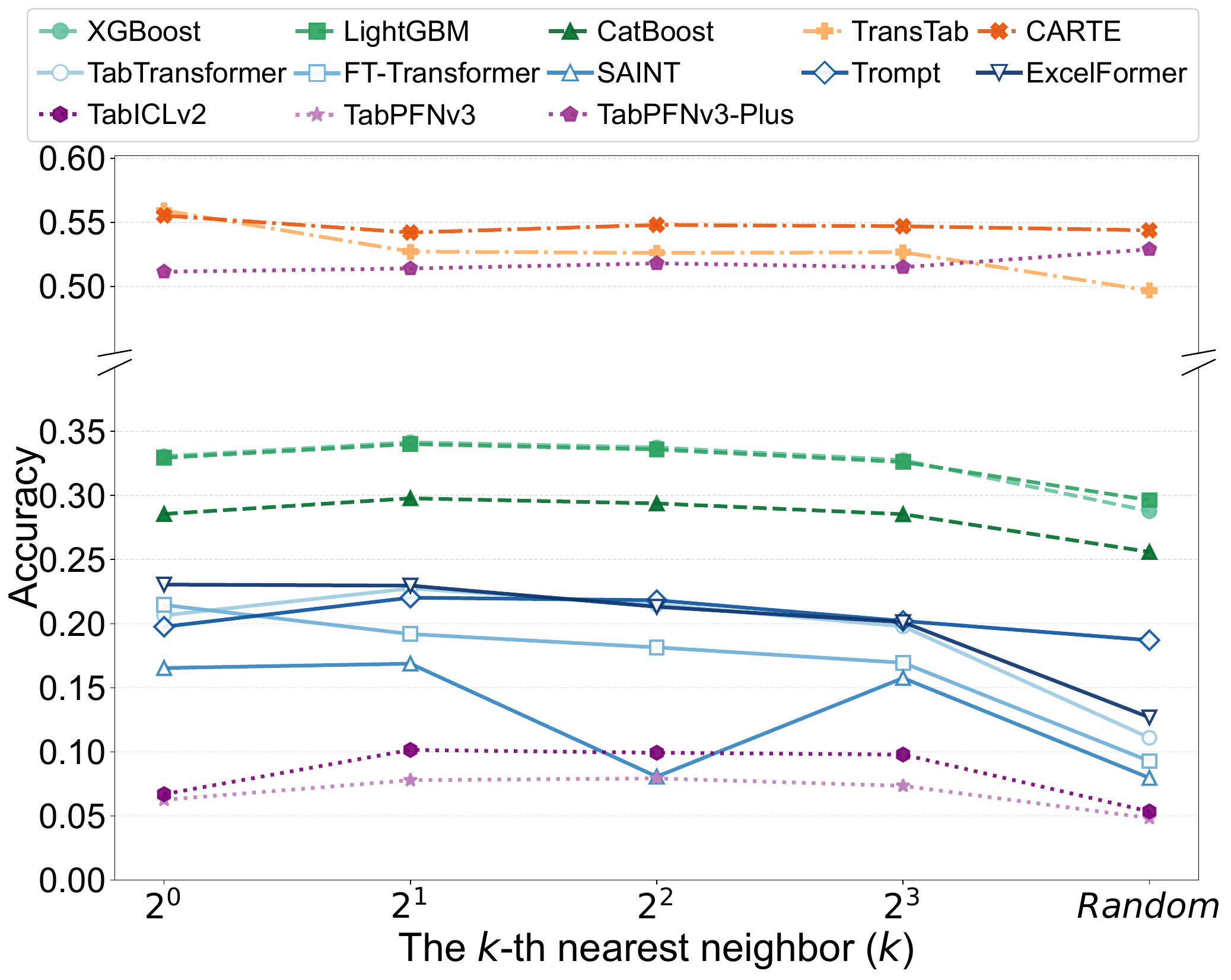} 
\caption{Impact of match quality on accuracy (join setting in AGBooks).}
\label{fig:exp_join_diff_k_agbooks}
\end{figure}

\section{Category-level Statistics of Auxiliary Strategy Gains}
\label{app:category_stats}

Table~\ref{tab:main_results} provides a category-level summary of the performance gains achieved by different auxiliary-information utilization strategies under the Union and Join relational settings. We group the evaluated methods into four categories---classical tree-based models, single-table tabular neural networks, transfer learning-based tabular models, and foundation models for tabular data. For each category, we aggregate the improvements obtained using three auxiliary-information utilization strategies, namely PT, DA, and FA, relative to training on the target table alone. To characterize both the consistency and magnitude of these improvements, we report (i) \textbf{Win\%}, the percentage of experimental configurations within each category that achieve a positive performance gain, and (ii) \textbf{Avg. Gain}, the average signed performance difference relative to the target-table-only baseline. For presentation, the win proportions and average performance differences are multiplied by 100; consequently, Win\% is expressed as a percentage, whereas Avg. Gain is expressed in percentage points. These statistics complement the main results by showing how consistently and substantially different method families benefit from auxiliary information across the two relational settings.

We make the following observations. \textbf{For the Union setting}, first, the pre-training-based transfer strategy (PT) provides the most consistent performance gains. Second, tree-based models benefit relatively consistently from both DA and FA. Third, CARTE achieves stable transfer improvements, as its graph-based representation facilitates cross-domain modeling, while its ensemble mechanism helps mitigate negative transfer. Finally, TransTab exhibits less stable transfer performance because it is more sensitive to structural and semantic discrepancies across tables; when source--target compatibility is limited, its pre-training--fine-tuning paradigm may overfit source-domain knowledge and lead to negative transfer. \textbf{For the Join setting}, first, FA yields the most substantial gains, indicating that auxiliary features can effectively enrich target representations, although the resulting benefit depends strongly on matching quality. Second, PT is also effective: TransTab and CARTE can partially accommodate source--target discrepancies, while the three datasets involve real-world entities with rich semantic priors, allowing the models to benefit from knowledge acquired during pre-training.

\end{document}